\documentclass[lettersize,journal]{IEEEtran}
\pdfoutput=1
\usepackage{amsmath,amsfonts}
\usepackage{algorithmic}
\usepackage{algorithm}
\usepackage{array}
\usepackage[caption=false,font=normalsize,labelfont=sf,textfont=sf]{subfig}
\usepackage{textcomp}
\usepackage{stfloats}
\usepackage{url}
\usepackage{verbatim}
\usepackage{graphicx}
\usepackage{cite}

\usepackage{times}
\usepackage{epsfig}
\usepackage{graphicx}
\usepackage{amsmath}
\usepackage{amssymb}
\usepackage{eso-pic}

\usepackage{lineno,hyperref}
\hypersetup{colorlinks=true,linkcolor=black}
\usepackage[numbers,sort&compress]{natbib}
\usepackage{ragged2e} 
\usepackage{booktabs,makecell, multirow, tabularx}

\begin{document}

%\title{Adaptive Instance-level Representation for Lifelong Person Re-Identification}
\title{Diverse Representations Embedding for Lifelong Person Re-Identification}

\author{Shiben Liu \href{https://orcid.org/0000-0001-9376-2562}{\includegraphics[scale=0.08]{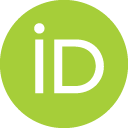}}, 
Huijie Fan*
\href{https://orcid.org/0000-0002-8548-861X}{\includegraphics[scale=0.08]{ORCIDiD_icon128x128.png}}, 
Qiang Wang \href{https://orcid.org/0000-0002-2018-1764}{\includegraphics[scale=0.08]{ORCIDiD_icon128x128.png}}, 
Xiai Chen \href{https://orcid.org/0000-0003-4756-3962}{\includegraphics[scale=0.08]{ORCIDiD_icon128x128.png}}, 
Zhi Han \href{https://orcid.org/0000-0002-8039-6679}{\includegraphics[scale=0.08]{ORCIDiD_icon128x128.png}}, 
Yandong Tang \href{https://orcid.org/0000-0003-3805-7654}{\includegraphics[scale=0.08]{ORCIDiD_icon128x128.png}}
\thanks{This work is supported by the National Natural Science Foundation of China (62273339, 61991413, U20A20200), and the Youth Innovation Promotion Association of Chinese Academy of Sciences (2019203). (\emph{Corresponding author: Huijie Fan})}
\thanks{Shiben Liu is with the State Key Laboratory of Robotics, Shenyang Institute of Automation, Chinese Academy of Sciences, Shenyang 110016,
China, and with the Institutes for Robotics and Intelligent Manufacturing,	Chinese Academy of Sciences, Shenyang 110169, China, and also with the
University of Chinese Academy of Sciences, Beijing 100049, China (e-mail: liushiben@sia.cn).
\par
Huijie Fan, Xiai Chen, Zhi Han, and Yandong Tang are with the State Key Laboratory of Robotics, Shenyang Institute of Automation, Chinese Academy of Sciences, Shenyang, 110016, China, and with the Institutes for Robotics and Intelligent Manufacturing, Chinese Academy of Sciences, Shenyang, 110016, China (e-mail: fanhuiie@sia.cn;  chenxiai@sia.cn; hanzhi@sia.cn; ytang@sia.cn).
\par 
Qiang Wang is with the Key Laboratory of Manufacturing Industrial Integrated Automation, Shenyang University, and with the State Key Laboratory of Robotics, Shenyang Institute of Automation, Chinese Academy of Sciences, Shenyang, 110016, China (e-mail: wangqiang@sia.cn). 
\par 
}% 
}

\markboth{}%
{Shell \MakeLowercase{\textit{et al.}}: A Sample Article Using IEEEtran.cls for IEEE Journals}


\maketitle

\begin{abstract}
Lifelong Person Re-Identification (LReID) aims to continuously learn from successive data streams, matching individuals across multiple cameras. The key challenge for LReID is how to effectively preserve old knowledge while incrementally learning new information, which is caused by task-level domain gaps and limited old task datasets. Existing methods based on CNN backbone are insufficient to explore the representation of each instance from different perspectives, limiting model performance on limited old task datasets and new task datasets. Unlike these methods, we propose a Diverse Representations Embedding (DRE) framework that first explores a pure transformer for LReID. The proposed DRE preserves old knowledge while adapting to new information based on instance-level and task-level layout. Concretely, an Adaptive Constraint Module (ACM) is proposed to implement integration and push away operations between multiple overlapping representations generated by transformer-based backbone, obtaining rich and discriminative representations for each instance to improve adaptive ability of LReID. Based on the processed diverse representations, we propose Knowledge Update (KU) and Knowledge Preservation (KP) strategies at the task-level layout by introducing the adjustment model and the learner model. KU strategy enhances the adaptive learning ability of learner models for new information under the adjustment model prior, and KP strategy preserves old knowledge operated by representation-level alignment and logit-level supervision in limited old task datasets while guaranteeing the adaptive learning information capacity of the LReID model. Extensive experiments were conducted on eleven Re-ID datasets, including five seen datasets for training in order-1 and order-2 orders and six unseen datasets for inference. Compared to state-of-the-art methods, our method achieves significantly improved performance in holistic, large-scale, and occluded datasets. \textbf{Our code will be available soon.}
\end{abstract}


\section{Introduction}
\begin{figure}[t]
	\centering
	\includegraphics[width=0.99\linewidth, height=0.33\textheight]{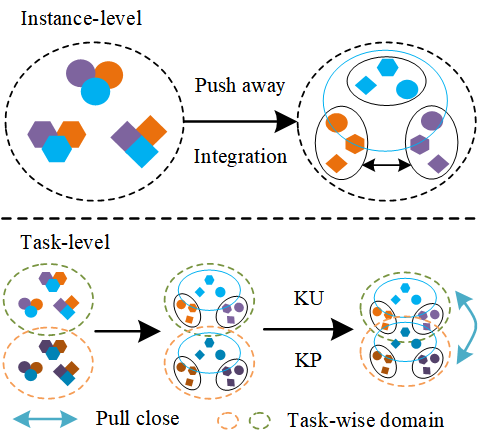}
	\caption{An illustration of our DRE. Each shape denotes an instance, and color indicates a different representation of the same instance. The blue circle indicates the primary embedding representation after integration. At the top, the generated diverse representation is implemented for integration and push operation. Integrated primary embedding representation (blue circle) presents rich body information. Multiple separate auxiliary embedding representations (orange and purple) maintain discrimination of each instance. At the bottom, based on instance-level, we introduce Knowledge Update (KU) and Knowledge Preservation (KP) strategies to improve model performance in preserving old knowledge and adapting to new information in sequential tasks.}
	\label{fig:Fig1}
\end{figure}
\IEEEPARstart{P}{erson} re-identification (ReID) aims to retrieve the same individuals across different camera views \cite{wei2021flexible, pan2023pose, huang2023reasoning, li2023clip}, which is crucial in applications, such as intelligent surveillance \cite{eom2021disentangled, lin2022learning, tan2022mhsa} and multi-camera collaborative tracking\cite{quach2021dyglip, yu2023region, hu2023temporal}. Most ReID methods assume that the training datasets can be accessed all at once, whereas ReID datasets are continuously incoming in practical applications. Lifelong learning is necessary for person re-identification in continuous datasets across the scene. \\
\indent Most ReID methods \cite{zhou2021attention, zhao2023content} learn robust features on specific distribution datasets through pre-trained networks. Unlike ReID, the main purpose of LReID is to facilitate model to efficiently preserve old knowledge while adapt to new information in sequential tasks. The rehearsal-based \cite{wu2021generalising, pu2021lifelong} methods reserve a memory buffer for storing limited instances from old tasks, and recall these instances when learning new tasks, which ignore model representation capabilities on limited old task datasets. A framework of learnable and consistent features \cite{huang2023learning} is proposed that provides complementary knowledge for diverse representations generation, improving the discrimination and adaptability of the LReID model. Inspired by knowledge distillation \cite{hinton2015distilling}, several methods \cite{ge2022lifelong, sun2022patch} in LReID incorporate distillation loss between teacher networks with prior knowledge of old tasks and student networks with adaptive learning of current tasks to mitigate task-level domain gaps. These methods all employ fixed-weight teacher networks, forcing the model to reduce its adaptability to new tasks. Yu \emph{et al.} \cite{yu2023lifelong} introduced a knowledge refresh scheme that updates the memory model with a smaller learning rate, which turns the memory model into a dynamic teacher. The above CNN-based methods insufficiently explore the discriminative and rich representations of each instance, limiting the performance of the LReID model in preserving old knowledge while adapting to new information. We consider that the CNN backbone (ResNet50, ResNet101) forces the introduction of multiple branches to obtain diverse representations that are independent and uncontrollable from each other in the LReID model.\\
\indent Recently, Vision Transformer (ViT) \cite{dosovitskiy2020image} and Data-efficient Image Transformers (DeiT) \cite{touvron2021training} have achieved significant success in image recognition \cite{xu2022multiple, zhang2023cross}, detection \cite{wang2022fpt, zhu2023transformer}, and restoration \cite{zhang2023lrt, wang2023uscformer}. In addition, TransReID \cite{he2021transreid} indicates that pure transformer is effective in feature extraction for ReID, because multi-head self-attention of the transformer frame captures long-range dependencies and drives the model to attend diverse human body parts.  DC-Former\cite{li2023dc} gets multiple diverse and compact embedding subspaces. Each embedding of these compact subspaces is more robust and discriminative to identify similar classes. Although the above methods have gained significant advantages in ReID, transformers still need to be specifically designed for LReID to preserve old knowledge while adapting new information. \\
\indent Side information embeddings \cite{he2021transreid} (such as camera and viewpoint-specific information) constantly change as the number of tasks increases in LReID, leading to bias in the LReID model. We introduce Maximum Embedding (ME) to replace side information embeddings, focusing on critical regions of the input sequence and reducing background interference. Meanwhile, the class token of transformer is transformed into a global representation of each instance in the ReID task. We introduce multiple class tokens embedding to learn multiple overlopping representation of each instance in the LReID task. Multiple class tokens and patches are concatenated in the first dimension and sent to the transformer for training on the seen dataset. Therefore, we construct a strong backbone based on a pure transformer in LReID, as an adjustment model and learner model structure.  \\
\indent Based on the above improved transformer backbone, we propose a diverse representation embedding that first explores the pure transformer for LReID, named DRE. Specifically, multiple overlopping representations are generated by multiple class tokens of transformer, as shown in Figure \ref{fig:Fig1} (instance-level). In this paper, we set up multiple overlopping representations. One is the primary embedding representation for learning rich body information, the other is multiple auxiliary embedding representations for providing discrimination of each instance. We propose an Adaptive Constraint Module (ACM) to implement integration and push away operations between multiple overlopping representations. Multiple auxiliary embedding representations minimize overlapping elements by orthogonal operation. We adaptively integrate multiple discriminative auxiliary embedding representations into primary embedding representations to enhance the representational capacity of the LReID model, which provides rich body information for classification. At the instance level, diverse representation generation presents richness and discrimination of each instance, remarkably improving feature learning and label matching processes for models with the same ID. Based on instance-level, we propose Knowledge Update (KU) and Knowledge Preservation (KP) strategies at the task-level layout by introducing the adjustment model and the learner model. KU strategy enhances the adaptive learning ability of learner models for new information under adjustment model prior. Because the knowledge update strategy is prone to adapt to new tasks, the knowledge preservation scheme is proposed to preserve old knowledge operated by representation-level alignment and logit-level supervision in limited old task datasets while guaranteeing the adaptive learning information capacity of the LReID model. The main contributions of this paper are summarized as follows:\\
\begin{itemize}	
	\item[$\bullet$] We propose a diverse representation embedding framework that first exploits the transformer-based backbone for LReID tasks. Maximum embedding Multiple class tokens are introduced to reduce bias and generate multiple representations of each instance.
	\item[$\bullet$] We design an adaptive constraint module to maintain rich primary embedding and discriminative auxiliary embedding representations in all seen domains, improving the adaptive capability of the LReID model.
	\item[$\bullet$] Based on diverse representations, knowledge update facilitate learning new information capability of the learner model, and knowledge preservation leverages representation-level alignment and logit-level supervision to preserve old knowledge while guaranteeing the learning information ability of the LReID model.
	\item[$\bullet$] Extensive experiments demonstrate that the proposed DRE outperforms state-of-the-art methods on both seen and unseen datasets. In general, our DRE outperforms performance with CNN-based methods.
\end{itemize} 
\section{Related work}
\subsection{Person Re-Identification}
\indent Research in person re-identification (ReID) make significant achievements over the years. These methods can be broadly divited into four kategories. First, CNN-bsed approaches \cite{xu2022learning,wang2022quality,zhang2023diverse,feng2023shape} extract discriminative features from pedestrian images to effectively address the spatial complexities. Cross-entropy loss \cite{zheng2017discriminatively} and triplet loss \cite{liu2017end} are commonly applied for training CNN to learn better representation. Second, transformer-based methods \cite{he2021transreid, jia2022learning, zhang2023pha, fan2023skip} make further progress  in ReID for their ability to capture long-range dependencies in feature maps. In these methods, self-attention mechnism is integrated into ReID to enhance the extraction of relevant information. Third, domain adaptation and transfer learning techniques \cite{wang2020exploiting, wang2021learning, liu2023discriminative, chen2023unsupervised} are also explored to improve model generalization, particularly when dealing with variations in lighting, viewpoints, and camera sources. Fourth, text-to-image methods \cite{chen2023towards, jiang2023cross, li2023clip} leverage textual descriptions and image data to improve recognition accuracy and model performance, enabling more effective matching of individuals in different contexts by combining text descriptions with image-based features.\\
\subsection{Lifelong Person Re-Identification}  
\indent Lifelong Person Re-Identification (LReID) faces a formidable challenge, aiming to address the evolving nature of person identification across various scenarios and domains. Some works \cite{pu2021lifelong, zhang2023spatial, zhang2023spatial, pu2023memorizing} are proposed to tackle the issue of adapting ReID models over time while retaining knowledge gained from previous experiences. Generally, Pu \emph{et al.} \cite{pu2021lifelong} proposed learnable knowledge graphs that adaptively facilitate the mutual exchange of new and old knowledge, thus achieving knowledge accumulation. Some works \cite{pu2022meta, ge2022lifelong, sun2022patch, huang2023learning} aim to extract rich and discriminative representation, mitigating the risk of knowledge forgetting. Pu \emph{et al.} \cite{pu2022meta} proposed meta-reconciliation normalization (MRN) for mining meta-knowledge shared across different domains. Meanwhile, ConRFL \cite{huang2023learning} maintains learnable and consistent features across all seen domains, which improves the discrimination and adaptation ability of the LReID model. In addition, some methods \cite{pu2021lifelong, yu2023lifelong, ge2022lifelong} mitigate catastrophic forgetting and enhance model accuracy by using rehearsal-based strategies with images stored from previous tasks.\\
\subsection{Diverse representation learning}  
\indent Representation learning methods employ specific modules or branches to mine multiple discriminative features of each instance, which minimize intra-class distance and maximize inter-class distance. It effectively improves model discrimination for identifying similar classes. DEEN \cite{zhang2023diverse} can effectively learn the informative representations by generating diverse embeddings subspace of each instance. DC-Former \cite{li2023dc} employs multiple class tokens in vision transformer to generate multiple representations, which increases the identity density of embedding space to help model improve its discrimination for identifying similar classes. Sun \emph{et al.} \cite{sun2022patch} employed adaptively-chosen patches to exploit body local information. Ge \emph{et al.} \cite{ge2022lifelong} generated pseudo-task features by a pseudo-task transformation module to complement the limited exemplars.\\
\indent These LReID methods employ CNN as a feature extractor, which insufficiently exploits rich and discriminative representations of each instance. We consider that diverse representations generated by multiple class tokens with transformer-specific structure play an important role in maintaining a trade-off between preserving old knowledge and adapting to new information. Thus, we first propose a transformer-based diversity representation embedding for lifelong person re-identification.
\begin{figure*}[!htbp]
	\centering 
	\includegraphics[width=0.96\linewidth, height=0.53 \textheight]{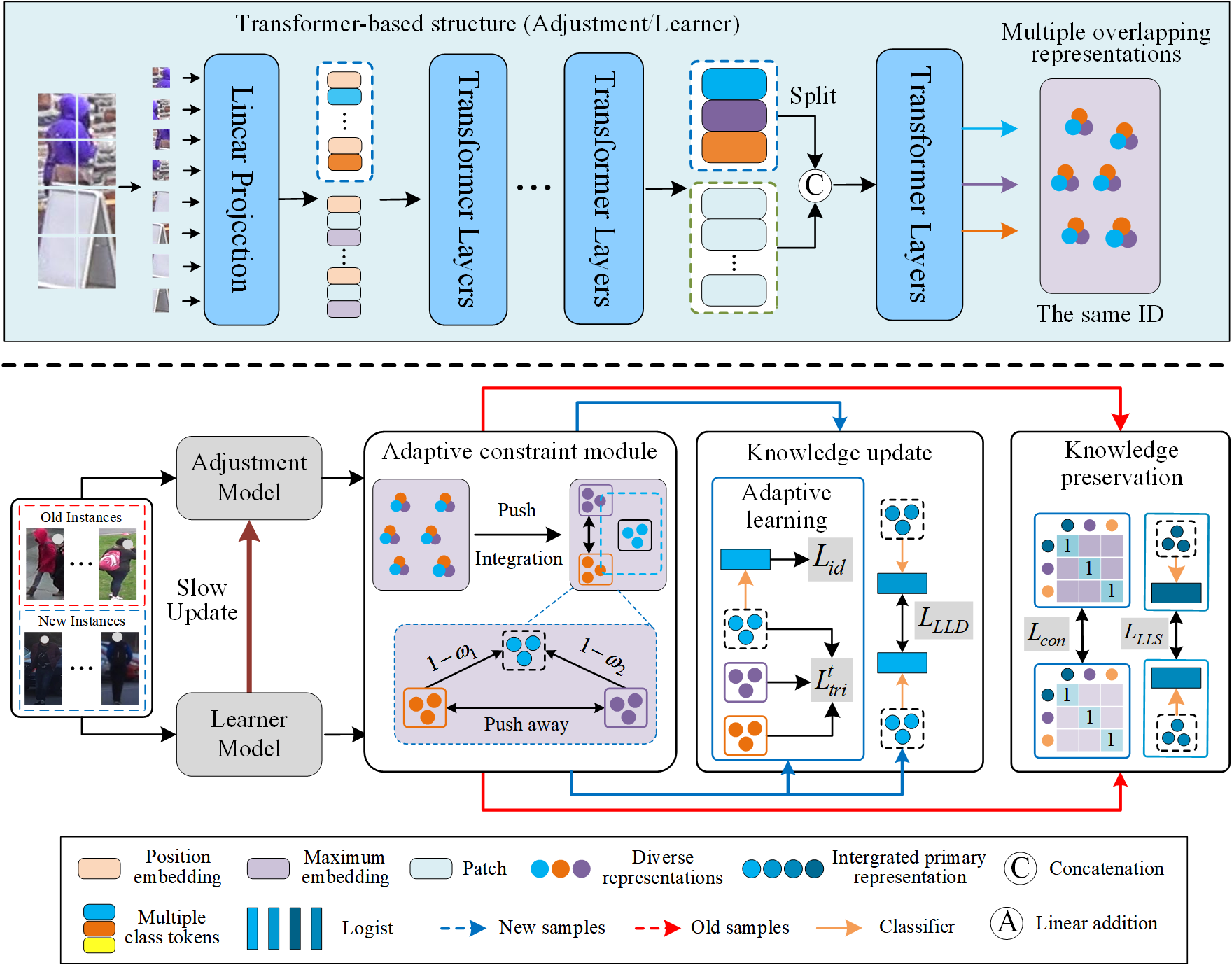}
	\caption{An overview of the proposed DRE for LReID. We first explore a pure transformer to LReID and equip it with an adjustment model and a learner model. The adaptive constraint module learns rich and discriminative representations. We then introduce knowledge update and knowledge preservation strategies based on diverse representations that collaboratively achieve knowledge interaction to preserve old knowledge while adapting to new information.} 
	\label{fig:fig2}
\end{figure*}
\section{Proposed Method}
\subsection{Overview}
\indent To preserve old knowledge while adapting new information in the face of limiting old task datasets and task-level domain gaps, we propose a Diverse Representation Embedding (DRE) framework based on improved transformer backbone for LReID. The input images are mapped into three representations using a transformer-based backbone. Three representations, including a primary embedding representation and multiple auxiliary embedding representations, are operated by Adaptive Constraint Module (ACM) to maintain richness and discrimination of each instance. Knowledge Update (KU) and Knowledge Preservation (KP) are proposed at the task-level layout by introducing the adjustment model and the learner model. We define the problem of lifelong person re-identification in this section. Specifically, continuous person re-identification datasets $E=\{E^t\}_{t=1}^T$ are collected in different environments. $E^t=\{E_{train}^t$ and $E_{test}^t\}$ indicates the training datasets and corresponding test datasets in t-th step, respectively. The training datasets $E_{train}$ are fed into the model in sequence for training. To store a limited number of instances from each old task dataset, we establish a memory buffer $M^t$. \\
\indent The DRE consists of an adjustment model $\Phi_{a}^t$ and a learner model $\Phi_{l}^{t}$ to extract diverse representations of each instance. The adjustment model maintains the priority of old tasks, and the learner model is responsible for adaptive learning of information for new instances. An adjustment model and a learner model both use improved transformer backbones. $\phi_{a}^t$ and $\phi_{l}^t$ serve as classifier heads for the adjustment and learner models, providing logits of each instance for recognition, as shown in Figure \ref{fig:fig2}. The outputs of old instances $x^o$ and new instances $x^n$ through adjustment model $\Phi_{a}^{t}$ and learner model $\Phi_{l}^{t}$ are as follows:
\begin{equation}
	\begin{aligned}
		&P_{l}^n, (A_{l}^n)^S = \Phi_{l}^{t}(x^n); &P_{a}^n, (A_{a}^n)^S = \Phi_{a}^{t}(x^n) \\
		&P_{l}^o, (A_{l}^o)^S = \Phi_{l}^{t}(x^o); &P_{a}^o, (A_{a}^o)^S = \Phi_{a}^{t}(x^o) \\
	\end{aligned}
\end{equation}
Where $P_{a}$ and $P_{l}$$\in\mathbb{R}^{B\times D}$ denote the primary embedding representations from the adjustment and learner models. $(A_{a})^S$ and $(A_{l})^S$$\in\mathbb{R}^{B\times D}$ are auxiliary embedding representations from the adjustment and learner models. $S$ is the number of auxiliary embedding representations.  
\subsection{Transformer-based structure}
\indent Currently, LReID methods based on CNN backbone force the introduction of multiple branches to obtain diverse representations that are independent and uncontrollable from each other in sequential task datasets. Multiple class tokens of transformer better fit our idea of diverse representations. Thus, we first propose a transformer-based backbone to generate diverse representations of each instance in LReID, as shown in Figure \ref{fig:fig2} (Transformer-based structure). Unlike the universal transformer structure, we introduce multiple class tokens and maximum embedding. For a given image $x$, we split $x$ into N fixed-size patches ${x_i|i=1,2,...,N}$, and then concatenate on the first dimension.
\begin{equation}
	Z_0 = [F(x_1); F(x_2); ... ; F(x_N)]
\end{equation}
where $Z_0$$\in$$\mathbb{R}^{B\times N\times D}$. $F$ is a linear projection mapping the patches to $D$ dimensions. [ ; ; ] denotes concatenation on the first dimensions. \\
\noindent \textbf{Maximum Embedding:} We further design maximum embedding (ME) to focus on critical regions of the input sequence and reduce background interference. ME utilizes a one-hot method \cite{gao2020pose} along the first dimension of the original input sequence $Z_0$ to query the index $\theta$ with the maximum value. The prominent patch embedding matrix is obtained by setting the value corresponding to the index $\theta$ to 1 and filling the other positions with 0. The embedding matrix can be directly mapped on $Z_0$ to obtain the prominent embedding region. Then the prominent embedding region $\underset{\theta}{arg max}Z_{0}^\theta$ is embedded to the original input sequence $Z_{0}$ by Hadamard product to obtain the maximum embedding $M_E$, and the maximum embedding is defined as:
\begin{equation}
	M_E= \underset{\theta}{arg max}Z_{0}^\theta \times Z_{0}
\end{equation}
\noindent \textbf{Multiple Class Tokens:} We introduce multiple class tokens embedded in transformer input sequence. Multiple class tokens and patches are concatenated on the first dimension and sent to the transformer for training on the dataset, which are converted into multiple overlapping representations. Multiple overlapping representations provide multiple high-level semantic information for each instance from different perspectives. The specific composition of transformer input sequence is as follows: 
% Similarly, each pedestrian has multiple attribute descriptions, such as long hair, height, girl, etc.
\begin{equation}
	Z_1 = [x_{cls}^0; ...; x_{cls}^{(S)}; M_E]
\end{equation}
Where $Z_1$$\in$$\mathbb{R}^{B\times(N+S+1)\times D}$. The input sequence $Z$ is expressed as:
\begin{equation}
	Z = Z_1 + P
\end{equation}
Where $P$$\in$$\mathbb{R}^{B\times (N+S+1)\times D}$ is position embedding.\\
\subsection{Adaptive constraint module}
\indent PatchKD \cite{sun2022patch} exploits local adaptively-chosen patches to pilot anti-forgetting distillation. However, when encountering large occluded regions, patches inevitably introduce non-body areas that weaken model performance in both old and new tasks. Therefore, we propose an adaptive learning module to enable multiple overlapping representations to retain discriminative abilities and rich body information of each instance. It plays a crucial role in preserving old knowledge while adapting to new tasks. As shown in Figure \ref{fig:fig2}.\\
\indent Specifically, the generated diverse representations are divided into a primary embedding representation and multiple auxiliary embedding representations. Multiple auxiliary embedding representations with discrimination are beneficial for improving the adaptive capacity of the LReID model. Therefore, we implement the minimization of overlapping elements among multiple auxiliary representations embedding by orthogonal. The orthogonal loss function is defined as:
\begin{equation}
	L_{ort} = \frac{1}{B}\frac{1}{S}\sum\limits_{i=1}^{B}\sum\limits_{j=1}^{S}(Cos<(A_l^n)^1,(A_l^n)^S> * I)
\end{equation}
Where $Cos<.,.>$ donates the cosine similarity, $I$ $\in\mathbb{R}^{S\times S}$ is a unit matrix, $(A_l^n)^S$ indicates auxiliary embedding representations from learner model in new instances, $S$ is the number of auxiliary embedding representations. \\
\indent The primary embedding representation is responsible for learning rich body information. Given that a single embedding representation inadequately captures rich body information, we adaptively integrate multiple discriminative auxiliary embedding representations into a primary embedding representation. This ensures that the model can better learn fine-grained and more informative representations for classification. The integrated primary embedding representation is formulated as:\\
\begin{equation}
	\begin{aligned}
		&\omega_S = Cos<P, (A_l^n)^S> \\
		\hat P = P + &(1-\omega_1)*(A_l^n)^1 + ... + (1-\omega_S)*(A_l^n)^S
	\end{aligned}
\end{equation}
Where, $\omega_S$$\in$[0,1] donates cosine similarity between the primary embedding representation and multiple auxiliary embedding representations. $P$, and $\hat P$ are primary embedding representation, and integrated primary embedding representation, respectively. Note that symbols $P$, $(A)^S$, and $\hat P$ use subscripts to indicate data flow direction in the following text. Their superscripts represent the flow of data ($n$ from new instances, $o$ from old instances), and the lower table represents the source of the model ($a$ from adaptation model, $l$ from learner model). Additionally, other symbols are defined according to specific contexts.\\
\subsection{Knowledge Update}
\indent In this section, we propose a Knowledge Update (KU) strategy that enhances the learner model's adaptive learning ability for new information under the adjustment model prior to new instances, including adaptive learning, logit-level distillation (LLD), and slow update.\\
\noindent \textbf{Adaptive learning:} We introduce cross-entropy loss and triplet loss to facilitate the adaptability of the learner model. The primary embedding representation of new instances are classified by classifier head of learner model is $\phi_{l}^t$. Cross-entropy loss is used to calculate identification loss. The identification loss $L_{id}$ is formulated as:
\begin{equation}
	L_{id} = -\frac{1}{B}\sum\limits_{i=1}^{B}(y^n)_ilog(\phi_{l}^t((\hat P_l^n)_i))
\end{equation}
Where $(y^n)$ is the label for new instances.\\
Meanwhile, we use metric learning to optimize the learner model on the primary and auxiliary embedding representations of new instances, and the triplet loss is defined as:
\begin{equation}
	L_{trip}^n = \frac{1}{B}\sum\limits_{i=1}^{B}max(d((f_n^a)_i, (f_n^p)_i)-d((f_n^a)_i, (f_n^n)_i) + m, 0)\\
\end{equation}
Where d(· , ·) denotes the euclidean distance. $m$ is set to 0.0 as default. ($f_n^a$,$f_n^p$,$f_n^n$) is a triplet of the anchor sample, positive sample and negative sample of [$\hat P_l^n$;$(A_l^n)^1$;...;$(A_l^n)^S$]. \\
\indent The base loss function for training the learner model on new instances is formulated as:
\begin{equation}
	L_{base} = L_{id} + L_{trip}^t + L_{ort}
\end{equation}
\noindent \textbf{Logit-level Distillation:} Base loss only considers adaptability of the learner model to new tasks, ignoring the consistency relationship between the adjustment model and the learner model. Guided by the feature distribution of the adjustment model, we further introduce logit-level distillation, focusing on mining consistency information from new samples. In particular, we employ a kullback-leibler divergence to minimize the logit-level distribution variance of primary embedding representation between the adjustment model and the learner model for new samples. $\phi_{a}^t$ and $\phi_{l}^t$ serve as classifier heads for the adjustment and learner models. Logit-level distillation loss is indicated as follows:
\begin{equation}
	L_{LLD} = \frac{1}{B}\sum\limits_{i=1}^{B}KL(\phi_{a}^t((\hat P_{a}^n)_i)/\tau)||\phi_{l}^t((\hat P_{l}^n)_i)/\tau))
\end{equation}
Where $KL(.||.)$ refers to a kullback-leibler divergence, and $\tau$ is a hyperparameter called temperature \cite{hinton2015distilling}.\\
\noindent \textbf{Slow Update:} The learner model tends to favor the distribution of new tasks, away from fixed-parameters adjustment model in terms of distribution. If we forcibly align the consistency of the adjustment and learner models to mitigate the risk of catastrophic forgetting, it will inevitably weaken adaptability of the learner model to new tasks. Therefore, we transform the frozen adjustment model into a dynamic adjustment model to improve the learner model's adaptability to new tasks. Specifically, we introduce an exponential moving average (EMA)  \cite{klinker2011exponential} to gradually update the adjustment model parameters using the learner model at each iteration step. The parameters for updating the adjustment model are formulated as:
\begin{equation}
	\Phi_{a}^{t} = k* \Phi_{a}^{t} + (1-k)*\Phi_{l}^{t}
\end{equation}
Where k is a constant and set to 0.996.
\subsection{Knowledge Preservation}
\indent Knowledge update effectively enhances the adaptability of the learner model to new tasks. However, it biases the distribution of the adjustment model toward new instances, ignoring the risk of forgetting old tasks. Therefore, we propose knowledge preservation (KP) strategy preserve old knowledge operated by representation-level alignment and logit-level supervision in limited old task datasets while guaranteeing the adaptive learning information capacity of the LReID model in terms of old instances from memory buffer $M^t$, including representation-level alignment (RLA) and logit-level supervision (LLS).\\
\noindent \textbf{Representation-level Alignment:} Since the LReID task is essentially a cross-camera matching problem, our goal is to learn significant representation rather than classification scores. We argue that the distribution between the adjustment model and the learner model under diverse representations of old instances should remain consistent to prevent forgetting old tasks. To achieve this goal, we first utilize triplet loss to optimize the primary and auxiliary embedding representations of the learner model obtained from old instances. The triplet loss is expressed as follows:
\begin{equation}
	L_{trip}^o = \frac{1}{B}\sum\limits_{i=1}^{B}max(d((f_o^a)_i, (f_o^p)_i)-d((f_o^a)_i, (f_o^n)_i) + m, 0)\\
\end{equation}
Where ($f_o^a$,$f_o^p$,$f_o^n$) is a triplet of anchor sample, positive sample and negative sample of [$\hat P_l^o$;$(A_l^o)^1$;...;$(A_l^o)^S$]. \\
\indent Then, we establish a consistent loss function for the primary and auxiliary embedding representations. Specifically, $F_{a}$=[$\hat P_a^o$;$(A_a^o)^1$;...; $(A_a^o)^S$]$\in$$\mathbb{R}^{B\times (S+1) \times D}$ and $F_{l}$=[$\hat P_l^o$;$(A_l^o)^1$;...;$(A_l^o)^S$] represent diverse representations concatenation of the adjustment model ($\Phi_{a}^{t}$) and the learner model ($\Phi_{l}^{t}$), respectively. 
We calculate cosine similarity between diverse representations of the adjustment and learner model and minimize the cosine similarity matrix by using the $L1$ norm to preserve old knowledge. Consistent loss $L_{con}$ is defined as:
\begin{equation}
	L_{con} = ||Cos<F_a, F_a>, Cos<F_l, F_l>||_1
\end{equation}
\indent Finally, the representation-level alignment loss is obtained through a linear combination of $L_{trip}^m$ and $L_{con}$. It is formulated as:
\begin{equation}
	L_{FLA} = L_{trip}^m + L_{con}
\end{equation}
\noindent \textbf{Logit-level Supervision:} Representation-level alignment only accounts for the relationship between the adjustment model and the learner model at the feature level while ignoring the interdependencies at the logit level. Therefore, we propose a logit-level supervision loss to bolster the preservation of old knowledge as a complement to representation-level alignment. Specifically, we leverage the logits from the primary embedding representation in the adjustment model as a supervisory signal to constrain the learner model. Logit-level supervision loss, named $L_{LLS}$, is defined as:
\begin{equation}
	L_{LLS} = -\frac{1}{B}\sum\limits_{i=1}^{B}\sigma(\phi_a^t((\hat P_a^o)_i))log(\sigma(\phi_l^t((\hat P_l^o)_i)))
\end{equation}
Where $\sigma$ is softmax function.\\
\indent The total loss function is formulated as:
\begin{equation}
	L =  L_{base} + L_{LLD} + L_{FLA} + L_{LLS}
\end{equation}
\begin{table}[htbp]
	\centering
	\renewcommand{\arraystretch}{1.3}
	\setlength{\tabcolsep}{7pt}
	\caption{Dataset statistics of the LReID benchmark. Since the sampling procedure results in the numbers of train IDs being all 500, the original numbers of IDs are listed for comparison. '-' denotes that the dataset is not used for training}\label{tab:Table1}
	\begin{tabular}{l|l|l|c|c}  %c:center r:right l:left
		\hline
		Type &Datasets &Scale &Train IDs &Test IDs \cr
		\hline
		\multirow{5}{*}{Seen}&
		Market \cite{zheng2015scalable} &large &500(751) &750 \cr
		&CUHK-SYSU \cite{xiao2016end} &mid &500(942) &2900 \cr
		&DukeMTMC \cite{ristani2016performance} &large &500(702) &1110 \cr
		&MSMT17$\_$V2 \cite{wei2018person} &large &500(1041) &3060 \cr
		&CUHK03 \cite{li2014deepreid} &mid &500(700) &700 \cr
		\hline
		\multirow{6}{*}{Unseen}&
		VIPeR \cite{gray2008viewpoint} &small &\makecell[c]{$-$} &316  \cr
		&GRID \cite{loy2010time} &small &\makecell[c]{$-$} &126 \cr
		&CUHK02 \cite{li2013locally} &mid &\makecell[c]{$-$} &239 \cr
		&Occ$\_$Duke \cite{miao2019pose} &large &\makecell[c]{$-$} &1100 \cr
		&Occ$\_$REID \cite{zhuo2018occluded} &mid &\makecell[c]{$-$} &200 \cr
		&PRID2011 \cite{hirzer2011person} &small &\makecell[c]{$-$} &649 \cr
		\hline
	\end{tabular}
\end{table}
\begin{table*}[!htbp]
	\centering
	\renewcommand{\arraystretch}{1.3}
	\setlength{\tabcolsep}{7pt}	
	\caption{Performance comparison with state-of-the-art methods for training order-1. Bold font indicates optimal values and red font is suboptimal values.}
	\label{tab:Table2}
	\begin{tabular}{c|cc|cc|cc|cc|cc|cc}
		\hline
		\multirow{2}{*}{Method}&
		\multicolumn{2}{c|}{Market}&\multicolumn{2}{c|}{CUHK-SYSU}&\multicolumn{2}{c|}{DukeMTMC}&\multicolumn{2}{c|}{MSMT17$\_$V2}&
		\multicolumn{2}{c|}{CUHK03}&\multicolumn{2}{c}{Seen-Avg}\\
		%\cmidrule(lr){2-4} \cmidrule(lr){5-7}\cmidrule(lr){8-10}
		\cline{2-13}
		&mAP&R-1&mAP&R-1&mAP&R-1&mAP&R-1&mAP&R-1 &$\overline{s}_{mAP}$ &$\overline{s}_{R-1}$ \cr
		\hline
		SPD\cite{tung2019similarity}&32.7&58.3&58.0&60.6&25.2&43.8&4.5 &13.1 &41.3 &43.4 &32.3 &43.9\\
		LwF\cite{li2017learning} &56.3 &77.1 &72.9 &75.1 &29.6 &46.5 &6.0 &16.6 &36.1 &37.5 &40.2 &50.6\\
		CRL\cite{zhao2021continual}&58.0 &78.2 &72.5 &75.1 &28.3 &45.2 &6.0 &15.8 &37.4 &39.8 &40.5 &50.8\\
		\hline
		AKA\cite{pu2021lifelong}&58.1 &77.4 &72.5 &74.8 &28.7 &45.2 &16.4 &37.6 &38.7 &40.4 &42.8 &55.1 \\
		PTKP\cite{ge2022lifelong}&64.4 &82.8 &79.8 &81.9 &45.6 &63.4 &10.4 &25.9 &42.5 &42.9 &48.5 &59.4\\
		PatchKD\cite{sun2022patch}&68.5 &\textcolor{red}{85.7} &75.6 &78.6 &33.8 &50.4 &6.5 &17.0 &34.1 &36.8 &43.7 &53.7\\
		KRKC\cite{yu2023lifelong}&54.0 &77.7 &83.4 &85.4 &48.9 &\textcolor{red}{65.5} &14.1 &33.7 &49.9 &50.4 &50.1 &62.5 \\
		ConRFL\cite{huang2023learning}&59.2 &78.3 &82.1 &84.3 &45.6 &61.8 &12.6 &30.4 &51.7 &53.8 &50.2 &61.7\\	
		\hline
		CODA\cite{smith2023coda}&53.6 &76.9 &75.7 &78.1 &48.6 &59.5 &13.2 &31.3 &47.2 &48.6 &47.7 &58.9\\
		CODA+ACM &54.1 &76.3 &78.6 &81.6 &50.1 &63.8 &15.3 &35.7 &51.3 &52.4 &49.9 &62.0\\
		\hline
		ResKUP &68.7 &84.6 &82.8 &84.5 &44.1 &60.9 &10.7 &26.1 &33.6 &34.3 &48.0 &58.1 \\
		DRE-KU &56.2 &76.7 &83.9 &84.2 &49.5 &62.4 &18.5 &39.6 &53.3 &56.5 &52.3 &63.9 \\
		FDRE-KUP &\textbf{72.4} &\textbf{87.7} &\textcolor{red}{85.5} &\textcolor{red}{86.1} &\textcolor{red}{50.6} &64.3 &\textcolor{red}{19.2} &\textcolor{red}{41.4} &\textcolor{red}{53.5} &\textcolor{red}{57.2} &\textcolor{red}{56.2} &\textcolor{red}{67.3}\\
		DRE&\textcolor{red}{69.4} &85.2 &\textbf{85.6} &\textbf{86.8} &\textbf{52.8} &\textbf{67.3} &\textbf{21.5} &\textbf{43.7} &\textbf{54.7} &\textbf{58.1} &\textbf{56.8} &\textbf{68.2} \\
		\hline
	\end{tabular}
\end{table*}
\begin{table*}[!htbp]
	\centering
	\renewcommand{\arraystretch}{1.3}
	\setlength{\tabcolsep}{7pt}
	\caption{Performance comparison with state-of-the-art methods for training order-2. Bold font indicates optimal values and red font is suboptimal values.}
	\label{tab:Table3}
	\begin{tabular}{c|cc|cc|cc|cc|cc|cc}
		\hline
		\multirow{2}{*}{Method}&
		\multicolumn{2}{c|}{DukeMTMC}&\multicolumn{2}{c|}{MSMT17$\_$V2}&\multicolumn{2}{c|}{Market}&\multicolumn{2}{c|}{CUHK-SYSU}& \multicolumn{2}{c|}{CUHK03}&\multicolumn{2}{c}{Seen-Avg}\\
		%\cmidrule(lr){2-4} \cmidrule(lr){5-7}\cmidrule(lr){8-10}
		\cline{2-13}
		&mAP &R-1 &mAP &R-1 &mAP &R-1 &mAP &R-1 &mAP &R-1 &$\overline{s}_{mAP}$ &$\overline{s}_{R-1}$ \cr 
		\hline
		SPD\cite{tung2019similarity} &28.5 &48.5 &3.7 &11.5 &32.3 &57.4 &62.1 &65.0 &43.0 &45.2 &33.9 &45.5 \\
		LwF\cite{li2017learning} &42.7 &61.7 &5.1 &14.3 &34.4 &58.6 &70.0 &73.0 &34.1 &34.1 &37.2 &48.4 \\
		CRL\cite{zhao2021continual} &43.5 &63.1 &4.8 &13.7 &35.0 &59.7 &70.0 &72.8 &34.5 &36.8 &37.6 &49.2 \\
		\hline
		AKA\cite{pu2021lifelong} &42.2 &60.1 &5.4 &15.1 &37.2 &59.8 &71.2 &73.9 &36.9 &37.9 &38.6 &49.4 \\
		PTKP\cite{ge2022lifelong} &54.8 &70.2 &10.3 &23.3 &59.4 &79.6 &80.9 &82.8 &41.6 &42.9 &49.4 &59.8\\
		PatchKD\cite{sun2022patch} &58.3 &74.1 &6.4 &17.4 &43.2 &67.4 &74.5 &76.9 &33.7 &34.8 &43.2 &54.1\\
		KRKC\cite{yu2023lifelong} &50.6 &65.6 &13.6 &27.4 &56.2 &77.4 &\textcolor{red}{83.5} &\textcolor{red}{85.9} &46.7 &46.6 &50.1 &61.0 \\
		ConRFL\cite{huang2023learning}&34.4 &51.3 &7.6 &20.1 &61.6 &80.4 &82.8 &85.1 &49.0 &50.1 &47.1 &57.4\\
		\hline
		CODA\cite{smith2023coda}&38.7 &56.6 &11.6 &24.5 &54.3 &75.1 &76.2 &75.8 &42.3 &41.7 &44.6 &54.7\\
		CODA+ACM &40.4 &59 &14.7 &29.0 &57.5 &78.6 &79.8 &80.7 &43.9 &44.9 &47.3 &58.4\\
		\hline
		ResKUP &59.4 &74.0 &11.9 &24.5 &46.8 &68.8 &80.1 &82.7 &39.6 &40.4 &47.5 &58.1 \\
		DRE-KU &49.2&65.9&16.3&31.8&58.2&76.7&82.2&84.6&47.4&48.5&50.7&61.5 \\
		FDRE-KUP&\textbf{62.3}&\textbf{76.4}&\textcolor{red}{18.1}&\textcolor{red}{33.6}&\textcolor{red}{63.7}&\textcolor{red}{81.4}&83.4&85.1&\textcolor{red}{50.3}&\textcolor{red}{51.9}&\textcolor{red}{55.6}&\textcolor{red}{65.7}\\
		DRE &\textcolor{red}{59.7}&\textcolor{red}{74.2}&\textbf{18.7} &\textbf{34.8} &\textbf{65.4} &\textbf{82.7} &\textbf{84.8} &\textbf{86.7} &\textbf{51.9} &\textbf{53.2} &\textbf{56.1} &\textbf{66.3} \\
		\hline
	\end{tabular}
\end{table*}

\section{Experiments}
\begin{table*}[!htbp]
	\centering
	\renewcommand{\arraystretch}{1.3}
	\setlength{\tabcolsep}{5pt}
	\caption{Training order-1 and training order-2 test on unseen datasets. Bold font indicates optimal values and red font is suboptimal values.}
	\label{tab:Table4}
	\begin{tabular}{c|cc|cc|cc|cc|cc|cc|cc}
		\hline
		\multirow{2}{*}{Method}&
		\multicolumn{2}{c|}{VIPeR}&\multicolumn{2}{c|}{GRID}&\multicolumn{2}{c|}{CUHK02}&\multicolumn{2}{c|}{Occ$\_$Duke}& \multicolumn{2}{c|}{Occ$\_$REID}&\multicolumn{2}{c|}{PRID2011}&\multicolumn{2}{c}{Unseen-Avg}\\
		\cline{2-15}
		&mAP &R-1 &mAP &R-1 &mAP &R-1 &mAP &R-1 &mAP &R-1 &mAP &R-1 &$\overline{s}_{mAP}$ &$\overline{s}_{R-1}$ \cr
		\hline
		AKA\cite{pu2021lifelong} &45.3&36.4&24.4&18.6&68.6&62.9&21.0&29.5&58.5&62.7&34.4&28.5&42.0&39.8 \\
		PTKP\cite{ge2022lifelong} &55.7&45.6&32.4&26.4&73.9&73&29.7&38.2&76.3&82.5&39.2&29&51.2&49.1 \\
		PatchKD\cite{sun2022patch} &48.4&39.2&28.1&23.8&72&69.5&24.5&32&65.8&70.6&31.6&24.7&45.1&43.3 \\
		KRKC\cite{huang2023learning} &54.2&43.7&36.2&28&81.6&83.2&26.7&35.6&76.9&82.6&40.8&31.9&52.7&50.8 \\
		\hline
		CODA \cite{smith2023coda}&47.8 &37.4 &26.2 &20.1 &71.4 &68.8 &22.1 &28.4 &67.1 &74.8 &32.6 &25.0 &44.5 &42.4\\
		CODA+ACM &49.6 &40.1 &28.5 &23.2 &75.4 &74.4 &25.7 &31.9 &73.9 &78.5 
		&34.3 &27.2 &47.9 &45.9\\
		\hline
		ResKUP &53.2&43.4&29.3&21.6&76.3&75.9&24.2&33.5&74.8&79.7&36.9&26.1&49.1&46.7 \\
		DRE-KU &\textcolor{red}{56.8}&\textcolor{red}{46.1}&36.9&27.5&80.1&79.9&28.8&37.3&78.2&83.8&41.9&32.4&53.8&51.2 \\
		DRE-KUP &54.5&45.2&\textbf{41.6}&\textbf{34.8}&\textcolor{red}{83.4}&\textcolor{red}{83.9}&\textcolor{red}{32.6}&\textbf{43.5}&\textcolor{red}{80.5}&\textcolor{red}{85.7}&\textcolor{red}{42.8}&\textcolor{red}{34.6}&\textcolor{red}{55.9}&\textcolor{red}{54.6} \\
		DRE &\textbf{57.2} &\textbf{47.6} &\textcolor{red}{40.1} &\textcolor{red}{33.6} &\textbf{83.9} &\textbf{84.5} &\textbf{33.2} &42.6 &\textbf{82.1} &\textbf{86.1} &\textbf{43.5} &\textbf{35.3} &\textbf{56.7} &\textbf{55.0} \\
		\hline
		\hline
		AKA\cite{pu2021lifelong} &44.3&35.1&23.8&18.6&71.8&69.2&17.3&20.8&61.2&68.3&29.6&22.1&41.3&39.0 \\
		PTKP\cite{ge2022lifelong} &52.5&43&27.1&18.4&77.9&77.8&33.1&40.8&76.9&81.4&37.3&28.0&50.8&48.2 \\
		PatchKD\cite{sun2022patch} &51.1&42.1&21.8&16.2&74.4&70.4&28.1&37.3&66.5&73.6&26.8&20.2&44.8&43.3 \\
		KRKC\cite{huang2023learning} &54.7&44.9&33.5&26.4&81.9&82.3&24.7&33.5&75.2&79.7&42.6&31.2&52.1&47.7 \\
		\hline
		CODA\cite{smith2023coda}&46.2 &37.6 &27.3 &19.8 &73.6 &66.8 &25.7 &35.6 &64.6 &71.4 &32.4 &25.7 &45.0 &42.9\\
		CODA+ACM &47.4 &39.8 &29.8 &22.6 &74.3 &69.3 &28.6 &38.4 &69.5 &76.5 &36.0 &29.2 &47.6 &46.0 \\
		\hline
		ResKUP &54.0&44.3&29.6&21.6&77.4&78.9&27.3&34.6&71.3&78.4&40.9&29.9&50.1&48.0 \\
		DRE-KU &55.1 &46.3 &34.9 &27.2 &81.3 &80.7 &29.8 &38.1 &77.4 &80.8 &43.4 &30.8&53.7&50.7 \\
		DRE-KUP &\textbf{58.1} &\textbf{49.8} &\textbf{38.5} &\textcolor{red}{29.6} &\textcolor{red}{82.5} &\textbf{83.6} &\textcolor{red}{36.8} &\textcolor{red}{43.2} &\textcolor{red}{80.5} &\textcolor{red}{85.3} &\textcolor{red}{43.7} &\textcolor{red}{35.2}&\textcolor{red}{56.7}&\textcolor{red}{54.5} \\
		DRE &\textcolor{red}{56.9} &\textcolor{red}{47.9} &\textcolor{red}{37.4} &\textbf{30.2} &\textbf{83.7} &\textcolor{red}{83.1} &\textbf{38.2} &\textbf{44.9} &\textbf{81.5} &\textbf{86.9} &\textbf{45.1} &\textbf{36.3} &\textbf{57.1} &\textbf{54.9} \\
		\hline
	\end{tabular}
\end{table*}
\begin{figure*}[t]
	\centering 
	\includegraphics[width=0.90\linewidth, height=0.18 \textheight]{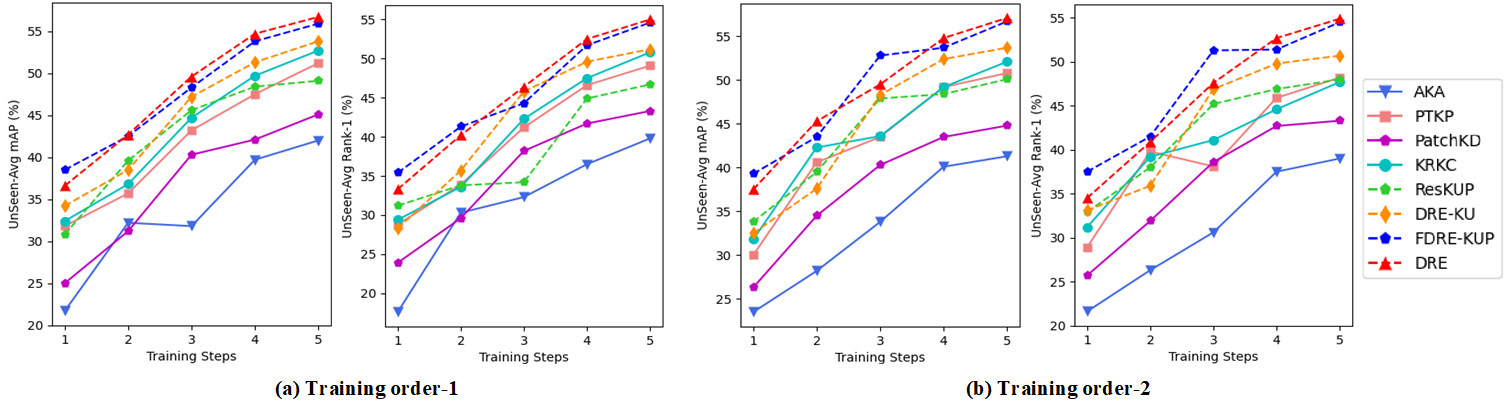}
	\caption{Generalization ability on unseen datasets during the training process.}
	\label{fig:fig3}
\end{figure*}
\subsection{Datasets and Evaluation Metrics}
\noindent \textbf{Datasets:} To evaluate the performance of our DRE for LReID tasks, we conduct extensive experiments in eleven datasets. Five consecutive datasets include Market \cite{zheng2015scalable}, CUHK-SYSU \cite{xiao2016end}, DukeMTMC \cite{ristani2016performance}, MSMT17$\_$V2 \cite{wei2018person} and CUHK03 \cite{li2014deepreid}, called seen datasets. We randomly sample 500 identities from each seen dataset for training. The others are unseen datasets used for generalization evaluation, consisting of VIPeR \cite{gray2008viewpoint}, GRID \cite{loy2010time},  CUHK02 \cite{li2013locally}, Occ$\_$Duke \cite{miao2019pose}, Occ$\_$REID \cite{zhuo2018occluded}, and PRID2011 \cite{hirzer2011person}. Unlike \cite{sun2022patch}, we exclude some uncommonly used datasets (CUHK01 \cite{li2013human}, SenseReID \cite{zhao2017spindle}, and I-LIDS \cite{zheng2009associating}) in ReID and introduce two occluded datasets (Occ$\_$Duke \cite{miao2019pose} and Occ$\_$REID \cite{zhuo2018occluded}) to enhance the diversity of unseen data, incorporating variations in shape, lighting conditions, and occlusion. During evaluation, all test datasets are merged into one benchmark. More detailed statistics for these datasets are provided in Table \ref{tab:Table1}.\\
\indent We investigate two representative training orders used in \cite{pu2021lifelong}. Training order-1 is Market$\to$CUHK-SYSU$\to$ DukeMTMC$\to$MSMT17$\_$V2$\to$CUHK03. Training order-2 represents DukeMTMC$\to$MSMT17$\_$V2$\to$Market$\to$ CUHK-SYSU$\to$ CUHK03.\\
\noindent \textbf{Implementation Details:} We employ VIT-B/16 \cite{dosovitskiy2020image}, initializing model weights from pre-training on ImageNet-21K \cite{he2021transreid}. All images are cropped to 256×128 pixels. Our DRE is trained for 60 epochs in each task using the SGD optimizer with a momentum of 0.9, and the batch size is set to 128. The initial learning rate is 0.032 with cosine learning rate decay.\\
\noindent \textbf{Evaluation Metrics:} We employ mean average precision (mAP) and Rank-1 accuracy (R-1) to evaluate performance on both seen and unseen datasets. We adopt the averag accuracy that indicates the average incremental accuracy of mAP and Rank-1 on both seen and unseen datasets. \\
\subsection{Comparison with State-of-the-Art methods}
\indent In this section, we compare our method to eight lifelong learning methods. SPD\cite{tung2019similarity}, LwF\cite{li2017learning}, and CRL\cite{zhao2021continual} are lifelong learning methods in natural settings, and LReID methods include AKA\cite{pu2021lifelong}, PTKP\cite{ge2022lifelong}, PatchKD\cite{sun2022patch}, KRKC\cite{huang2023learning}, and ConRFL\cite{huang2023learning}. 
To better validate the generalization of our method, we introduce a transformer-based CODA \cite{smith2023coda} method as the comparison method. The CODA+ACM method introduces our Adaptive Constraint Module (ACM) based on the CODA method. Additionally, variants of our approach are ResKUP, DRE-KU and FDRE-KUP. Compared to DRE, ResKUP uses ResNet50 as the feature extractor and outputs only one representation; DRE-KU does not incorporate a knowledge preservation strategy; FDRE-KUP freezes adjustment model parameters. Table \ref{tab:Table2} and Table \ref{tab:Table3} show the results of the comparison methods trained on the seen datasets with training order-1 and training order-2, respectively. Table \ref{tab:Table3} indicates the performance of the comparison methods tested on the unseen domain for training order-1 and order-2.\\
\noindent \textbf{Performance on Seen Datasets:} From Table \ref{tab:Table2}, it is evident that our DRE significantly outperforms comparison methods, with an average incremental gain of +6.6\% $\overline{s}_{mAP}$ and +6.5\% $\overline{s}_{R-1}$. On the large-scale dataset MSMT17$\_$V2, our DRE prominently improve performance by more than 1.4\% mAP and 6.1\% Rank-1. From Table \ref{tab:Table3}, the average improvement compared to other methods is 3.1\% and 2.1\% in terms of mAP and Rank-1, respectively. For first and last tasks, the average improvement over other methods is 3.1\% mAP and 0.1\% Rank-1, and 2.9\% mAP and 3.1\% Rank-1, respectively. \\
\indent To further validate the generalization of our method, we introduce transformer-based CODA method. From Table \ref{tab:Table2}-\ref{tab:Table4}, CODA method limits the performance of the model while preserving old knowledge. CODA+ACM method outperforms CODA methods in mAP and Rank-1 metrics, especially in occluded datasets (Occ$\_$Duke and Occ$\_$REID) and large scale datasets (MSMT17$\_$V2). This indicates that our Adaptive Constraint Module (ACM) can effectively improve the adaptive capability of the model, benefiting from rich and discriminative representations of each instance. \\
\indent To further validate the effectiveness of our method, we set up some variants, including ResKUP, DRE-KU, DRE-KUP. From Table \ref{tab:Table2} and Table \ref{tab:Table3}, ResKUP achieves poor mAP and Rank-1 performance. Because ResNet50 employs a branch to generate a representation of each instance, our Adaptive Constraint Model (ACM), Knowledge Update (KU) and Knowledge Preservation (KP) strategies, are not fully utilized in ResKUP. DRE-KU is superior to non-rehearsal methods (AKA, PatchKD, ConRFL), as it benefits from our KU and KP strategies based on diverse representations. DRE-KUP can effectively alleviate catastrophic forgetting problems, but its adaptive capacity is limited compared to DRE, due to the freezing of adaptive model parameters. In general, our DRE significantly improves performance in large-scale datasets to preserve old knowledge whlie adapting new information. \\
\noindent \textbf{Generalization Ability on Unseen Datasets:} In Table \ref{tab:Table4}, our DRE achieves a more stable result over six unseen datasets for training order-1 and training order-2. This demonstrates that our method allows smooth transfer and robust generalization across various unseen datasets. Compared to other methods, our DRE is more friendly to occlusion datasets (Occ$\_$Duke and Occ$\_$REID). We argue that ACM generates rich and discriminative representations to guide a dynamic balance between anti-forgetting and adapting to new tasks. Figure \ref{fig:fig3} illustrates that our method achieves optimal performance in terms of mAP and Rank-1 metrics for all unseen datasets during the training process. DRE-KU without using KP strategy, limiting the generalization capability of the LReID model. In contrast, our DRE consistently enhances generalization capabilities over time.\\
\noindent \textbf{Effectiveness of Transformer as a Feature Extrator:} We find that introducing diverse representations performs worse than using only one representation separately when employing ResNet50 as the feature extractor. Therefore, ResKUP uses only an embedding representation. Compared to ResKUP, DRE achieves significant performance improvement with an 8.6\% increase in mAP and an 8.2\% increase in Rank-1 for Seen-Avg, as shown in Table \ref{tab:Table3}. We attribute these improvements to ACM, KU and KP strategies, which generate rich and discriminative representations to interact knowledge between the adjustment model and the learner model. To further validate the effectiveness of the method, we compare the effects of different transformer backbones in the table below, such as DeiT-Base, DeiT-Small, and ViT-Small, as shown in Table \ref{tab:Table5}. ViT-Base and DeiT-Base are very close to mAP and Rank-1 metrics. The number of ViT-Small and DeiT-Small parameters decreased, resulting in lower mAP and Rank-1 indicators. The table shows that our method achieves effective performance on the transformer backbone. In summary, our proposed transformer-based DRE significantly improves performance on both old and new tasks over an extended period of time.\\
\begin{table}[t]
	\centering
	\renewcommand{\arraystretch}{1.3}\setlength{\tabcolsep}{10pt}
	\caption{Performance of different transformer-based backbone in training order-1.}
	\label{tab:Table5}
	\begin{tabular}{c|c|c|c|c}
		\hline
		\multirow{2}{*}{Backbone}&\multicolumn{2}{c|}{Seen$\_$Avg}&\multicolumn{2}{c}{Unseen$\_$Avg}\\
		%\cmidrule(lr){2-4} \cmidrule(lr){5-7}\cmidrule(lr){8-10}
		\cline{2-5}
		&$\overline{s}_{mAP}$&$\overline{s}_{R-1}$ &$\overline{s}_{mAP}$&$\overline{s}_{R-1}$ \cr \hline
		DeiT-Small&48.9&61.0&47.0&45.6  \\
		DeiT-Base&56.4&68.4&56.5&55.5  \\ 
		ViT-Small&48.3&60.3&48.4&48.0  \\
		ViT-Base&56.8&68.2&56.7&55.0  \\ 
		\hline
	\end{tabular}
\end{table}
\begin{table}[t]
	\centering
	\renewcommand{\arraystretch}{1.3}
	\setlength{\tabcolsep}{3pt}
	\caption{Ablation studies on the number of auxiliary embedding representation for ACM in training order-1.}
	\label{tab:Table6}
	\begin{tabular}{c|ccc|c|c|c|c}
		\hline
		\multirow{2}{*}{Primary}&\multicolumn{3}{c|}{Auxiliary}&\multicolumn{2}{c|}{Seen$\_$Avg}&\multicolumn{2}{c}{Unseen$\_$Avg}\\
		%\cmidrule(lr){2-4} \cmidrule(lr){5-7}\cmidrule(lr){8-10}
		\cline{2-8}
		&1&2&3&$\overline{s}_{mAP}$&$\overline{s}_{R-1}$ &$\overline{s}_{mAP}$&$\overline{s}_{R-1}$ \cr \hline
		$\surd$&&&&52.8&64.2&53.4&51.5  \\
		$\surd$&$\surd$&&&54.3&64.4&54.8&52.7  \\ 
		$\surd$&$\surd$&$\surd$&&\textbf{56.8}&68.2&\textbf{56.7}&\textbf{55.0}  \\
		$\surd$&$\surd$&$\surd$&$\surd$&56.4&\textbf{68.7}&56.0&54.6  \\ 
		\hline
	\end{tabular}
\end{table}
\begin{table}[!htbp]
	\centering
	\renewcommand{\arraystretch}{1.3}
	\setlength{\tabcolsep}{4pt}
	\caption{Performance of individual components for our DRE in training order-1. }
	\label{tab:Table7}
	\begin{tabular}{ccc|c|c|c|c}
		\hline
		\multirow{2}{*}{LLD}&\multirow{2}{*}{RLA}&\multirow{2}{*}{LLS}&\multicolumn{2}{c|}{Seen$\_$Avg}&\multicolumn{2}{c}{Unseen$\_$Avg}\\
		%\cmidrule(lr){2-4} \cmidrule(lr){5-7}\cmidrule(lr){8-10}
		\cline{4-7}
		&&&$\overline{s}_{mAP}$&$\overline{s}_{R-1}$&$\overline{s}_{mAP}$&$\overline{s}_{R-1}$ \cr \hline
		&&&49.4 &61.8 &50.4 &48.9 \\
		$\surd$ &&&52.3 &63.9 &53.8 &51.2 \\
		\hline
		$\surd$&$\surd$&&56.2 &67.6 &56.2 &54.5  \\
		$\surd$&&$\surd$&55.8 &66.1 &55.4 &53.3  \\
		&$\surd$&&54.8 &65.7 &53.2 &52.5  \\
		&$\surd$&$\surd$&55.3 &65.3 &55.8 &54.1  \\
		$\surd$ &$\surd$&$\surd$&\textbf{56.8} &\textbf{68.2} &\textbf{56.7}&\textbf{55.0}  \\ 
		\hline
	\end{tabular}
\end{table}
\subsection{Ablation Studies}
\noindent \textbf{Effectiveness on the number of auxiliary embedding representations for ACM:} The proposed ACM adaptively explores rich and discriminative representations. Here, we study the suitability of multiple auxiliary embedding representations for ACM. As shown in Table \ref{tab:Table6}, when the number of auxiliary embedding representations increases from 0 to 2, it generates rich and discriminative representations that generate multiple instances for the same classes to facilitate model performance improvement on both seen and unseen datasets. Therefore, ACM incorporates a primary embedding representations and two auxiliary embedding representations, achieving the best performance while striking a trade-off between metrics and complexity on both seen and unseen datasets. Therefore, the number of auxiliary embedding representations $S$ is set to 2.\\
\noindent \textbf{Performance of Individual Component:} To better evaluate the contribution of each component to DRE, we conduct some ablation studies on seen and unseen datasets. As shown in Table \ref{tab:Table7}, we can observe that the performance is unsatisfactory with only LLD, indicating that the performance of the knowledge update strategy reaches a limit in mitigating catastrophic forgetting. However, after introducing the RLA and LLS, the model performance is significantly improved, effectively achieving a trade-off between retaining old knowledge and adapting to new information. By integrating KU, KP, and ACM into an end-to-end LReID model, our proposed DRE achieves impressive performance improvements. 
\section{Conclusions}
\indent In this paper, we propose a diverse representation embedding framework that first exploits a pure transformer backbone to preserve old knowledge while adapting new information. First, the adaptive learning module facilitates diverse representations to maintain rich and discriminative body information for each instance. Then, we explore knowledge update and knowledge preservation strategies that collectively achieve knowledge interaction to mitigate catastrophic forgetting and adapt to new tasks over an extended period of time, and sufficiently explore the diverse representations of each instance based on transformer backbone. Extensive experiments demonstrate the superiority of our method in comparison with state-of-the-art LReID methods. Moreover, we conduct experiments to analyze the impact of the knowledge preservation strategy and the frozen adjustment model in our DRE, laying the foundation for future research.
%\section{Limitations And Future Works}
\bibliographystyle{unsrt}
\bibliography{document.bib}

\vfill

\end{document}